%% file: paper.tex
\documentclass[sigconf,natbib=true]{acmart}

\AtBeginDocument{%
  \providecommand\BibTeX{{%
    \normalfont B\kern-0.5em{\scshape i\kern-0.25em b}\kern-0.8em\TeX}}}

\setcopyright{acmcopyright}
\copyrightyear{2022}
\acmYear{2022}
\acmDOI{XXXXXXX.XXXXXXX}

\acmConference[]{x}{y}{z}
\acmPrice{15.00}
\acmISBN{978-1-4503-XXXX-X/18/06}

\usepackage[spacesep,definevectors]{easyvector}
\usepackage{amsmath}
\usepackage{mathtools}
\usepackage{xspace}
\usepackage{multirow}
\usepackage{algpseudocode}
\usepackage{algorithm}
\DeclareMathOperator*{\argmax}{argmax}

\begin{document}

\title{\FGETHR{}: Fine-Grained Named Entity Typing via Refinement in Hyperbolic Space}

\author{
	Muhammad Asif Ali,\textsuperscript{\rm 1}
	Yifang Sun,\textsuperscript{\rm 1}
	Bing Li,\textsuperscript{\rm 1}
	Wei Wang,\textsuperscript{\rm 2}\\
	\textsuperscript{\rm 1}School of Computer Science and Engineering, UNSW, Australia\\
	\textsuperscript{\rm 2}The Hong Kong university of Science and Technology, China\\
	\{muhammadasif.ali, bing.li, yifangs\}@unsw.edu.au, \{weiwcs\}@ust.hk\\
}





\newcommand{\revise}[1]{{\color{black}{#1}}}
\newcommand{\inv}[1]{#1^{-1}} 
\newcommand{\norm}[1]{\| #1 \|} 
\newcommand{\inprod}[2]{\langle #1, #2\rangle} 
\newcommand{\outprod}[2]{#1 \otime #2} 
\newcommand{\Relu}{\mathsf{ReLU}} 

\providecommand{\keywords}[1]{\textbf{\textit{Keywords ---}} #1}

\newcommand{\largevec}[1]{\overrightarrow{#1}}
\newcommand\cev[1]{\overleftarrow{#1}}
\renewcommand{\shortauthors}{Ali, et al.,}

\newcommand{\FGNET}{\text{FG-NET}}
\newcommand{\FGETHR}{\text{FGNET-RH}}
\newcommand{\fixme}[1]{\footnote{\textbf{FIXME!!!} #1}} 

\begin{abstract}
Fine-Grained Named Entity Typing (\FGNET{}) aims at classifying the entity mentions 
into a wide range of entity types (usually hundreds) depending upon the context. 
While distant supervision is the most common way to acquire supervised training data, 
it brings in label noise, as it assigns type labels to the entity mentions 
irrespective of mentions' context. In attempts to deal with the label noise, 
leading research on the \FGNET{} assumes that the fine-grained entity typing data 
possesses a euclidean nature, which restraints the ability of the existing models 
in combating the label noise. Given the fact that the fine-grained type hierarchy 
exhibits a hierarchical structure, it makes hyperbolic space a natural choice to model 
the \FGNET{} data. In this research, we propose \FGETHR{}, a novel framework that 
benefits from the hyperbolic geometry in combination with the graph structures to 
perform entity typing in a performance-enhanced fashion. \FGETHR{} initially uses 
LSTM networks to encode the mention in relation with its context, later it forms a 
graph to distill/refine the mention's encodings in the hyperbolic space. Finally, 
the refined mention encoding is used for entity typing. Experimentation using 
different benchmark datasets shows that \FGETHR{} improves the performance on \FGNET{} 
by up to 3.5\% in terms of strict accuracy.
\end{abstract}




\maketitle

\input{Chapter1}

\input{Chapter2}

\input{Chapter3}

\input{Chapter4}


\bibliographystyle{ACM-Reference-Format}
\bibliography{sample-base}

\end{document}

%% file: Chapter1.tex
\section{Introduction}
\label{intro}

Named Entity Typing (NET) is a fundamental operation in natural language processing, it aims at 
assigning discrete type labels to the entity mentions in the text. It has immense applications, 
including: knowledge base construction \cite{DBLP:conf/kdd/0001GHHLMSSZ14}; information 
retrieval \cite{lao2010relational}; question answering \cite{ravichandran2002learning}; 
relation extraction \cite{yaghoobzadeh2016noise} etc. Traditional NET systems work with only 
a coarse set of type labels, e.g., organization, person, location, etc., which severely limit 
their potential in the down-streaming tasks. In recent past, the idea of NET is extended to 
Fine-Grained Named Entity Typing (\FGNET{}) that assigns a wide range of correlated entity 
types to the entity mentions~\cite{DBLP:conf/aaai/LingW12}. Compared to NET, the \FGNET{} 
has shown a remarkable improvement in the sub-sequent applications. For example, 
Ling and Weld,~\cite{DBLP:conf/aaai/LingW12} showed that \FGNET{} can boost the 
performance of the relation extraction by 93\%. 

\begin{figure*}[!t]
    \centering
    \captionsetup{justification=centering}
    \resizebox{0.97\linewidth}{!}{
        \includegraphics[]{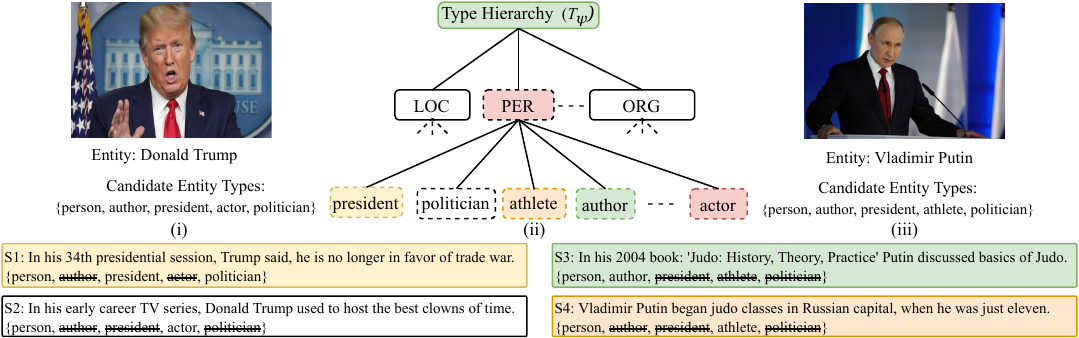}} 
    \caption{\FGNET{} training data acquired by distant supervision.~\revise{For examples S1:S4, we provide the fine-grained labels acquired by the distant supervision, with erroneous labels struck-through.}}
    \label{fig:noisy_data}
\end{figure*}

\FGNET{} encompasses hundreds of correlated entity types with little contextual differences, 
which makes it labour-intensive and error-prone to acquire manually labeled training data. 
Therefore, distant supervision is widely used to acquire training data for this task. 
Distant supervision relies on: \textit{(i)} automated routines to detect the entity mention, 
and \textit{(ii)} using type-hierarchy from existing knowledge-bases, e.g.,
Probase~\cite{DBLP:conf/sigmod/WuLWZ12}, to assign type labels to the entity mention. 
However, it assigns type-labels to the entity mention
irrespective of the mention's context, which results in
label noise~\cite{DBLP:conf/kdd/RenHQVJH16}. Examples in this regard are shown in 
Figure~\ref{fig:noisy_data}, where the distant supervision assigns labels: 
\{person, author, president, actor, politician\} to the entity mention: 
\emph{``Donald Trump"}, whereas, from contextual perspective, it should be labeled as: 
\{person, president, politician\} in S1, and \{person, actor\} in S2. Likewise, 
the entity mention: \emph{``Vladimir Putin"} should be labeled as: \{person, author\}  
and \{person, athlete\} in S3 and S4 respectively. This label noise in-turn propagates 
in the model learning and severely effects/limits the end-performance of the \FGNET{}
systems.

Earlier research on \FGNET{} either ignored the label
noise~\cite{DBLP:conf/aaai/LingW12}, or applied some heuristics to prune the
noisy labels~\cite{DBLP:journals/corr/GillickLGKH14}. Ren et
al.,~\cite{DBLP:conf/emnlp/RenHQHJH16} bifurcated the training data into
clean and noisy data samples, and used different set of loss functions to model them. 
However, the modeling heuristics proposed by these models are not able to cope with the 
label noise, which limits the end-performance of the \FGNET{} systems relying on distant 
supervision. We, moreover, observe that these models are designed assuming a euclidean 
nature of the problem, which is inappropriate for \FGNET{}, as the fine-grained type 
hierarchy exhibit a hierarchical structure. Given that it is not possible to embed 
hierarchies in euclidean space~\cite{DBLP:conf/nips/NickelK17}, this assumption, in 
turn limits the ability of the existing models to: \textit{(i)} effectively represent 
\FGNET{} data, \textit{(ii)} cater label noise, and \textit{(iii)} perform \FGNET{} 
classification task in a robust way. 

\begin{figure}[!]
	\centering
	\captionsetup{justification=centering}
	\resizebox{0.97\linewidth}{!}{
		\includegraphics[]{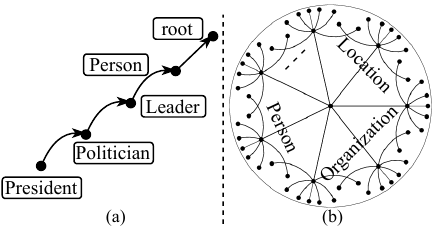}}
	\caption{ (a) Illustration of how the entity type \textit{``President"} shares the 
		context of the entity type \textit{``Politician"} which in turn shares the context 
		of the entity-type \textit{``Leader"} and so on; (b) Embedding \FGNET{} data in 2-D 
		Poincaré Ball, where each disjoint type~\revise{may potentially be} embedded along a different direction}
	\label{fig:HYP_EMB}
\end{figure}

The inherent advantage of hyperbolic geometry to embed hierarchies is well-established 
in literature. It enforces the items on the top of the hierarchy to be placed close to 
the origin, and the items down in the hierarchy near infinity. This enables the embedding 
norm to cater to the depth in the hierarchy, and the distance between embeddings represent 
the similarity between the items. Thus the items sharing a parent node are close to each 
other in the embeddings space. This makes the hyperbolic space a perfect paradigm for 
embedding the distantly supervised \FGNET{} data, as it explicitly allows label-smoothing 
by sharing the contextual information across noisy entity mentions corresponding to the 
same type hierarchy, as shown in Figure~\ref{fig:HYP_EMB} (b), for a 2D Poincar\'e Ball.
For example, given the type hierarchy: \emph{``Person"} $\leftarrow$ \emph{``Leader"} 
$\leftarrow$ \emph{``Politician"} $\leftarrow$ \emph{``President"}, the hyperbolic 
embeddings, on contrary to the euclidean embeddings, offer a perfect geometry for the 
entity type \emph{``President"} to share and augments the context of \emph{``Politician"}, 
which in turn adds to the context of \emph{``Leader"} and \emph{``Person"} etc., shown in 
Figure~\ref{fig:HYP_EMB} (a). We hypothesize that such hierarchically-organized contextually 
similar neighbours provide a robust platform for the end task, i.e., \FGNET{} over 
distantly supervised data, also discussed in detail in the section~\ref{subec:effectiveness}.

Nevertheless, we propose Fine-Grained Entity Typing with Refinement in Hyperbolic space 
(\FGETHR{}), shown in Figure~\ref{fig:proposed_approach}. \FGETHR{} follows a two-stage 
process, stage-I: encode the mention along with its context using multiple LSTM networks, 
stage-II: form a graph to refine mention's encoding from stage-I by sharing contextual 
information in the hyperbolic space. In order to maximize the benefits of using the 
hyperbolic geometry in combination with the graph structure, \FGETHR{} maps the mention 
encodings (from stage-I) to the hyperbolic space. And, performs all the operations: 
linear transformation, type-specific
contextual aggregation etc., in the hyperbolic space, required for appropriate additive 
context-sharing along the type hierarchy to smoothen the noisy type-labels prior to the 
entity typing. The major contributions of \FGETHR{} are enlisted as follows:

\begin{enumerate}
    
    \item \FGETHR{} accommodates the benefits of: the graph structures and the hyperbolic 
    geometry to perform fine-grained entity typing over distantly supervised noisy  data in a robust fashion. 
    
    \item \FGETHR{} explicitly allows label-smoothing over the noisy training data by using
     graphs to combine the type-specific contextual information along the type-hierarchy in the hyperbolic space.
    
    \item Experimentation using two models of the hyperbolic space, i.e., the Hyperboloid 
    and the Poincar\'e-Ball, shows that \FGETHR{} outperforms the existing research by up 
    to 3.5\% in terms of strict accuracy.
\end{enumerate}

%% file: Chapter2.tex
\section{Related Work}
\label{rl_work}
Existing research on \FGNET{} can be bifurcated into two major categories: \textit{(i)} 
traditional feature-based systems, and \textit{(ii)} embedding models. 

Traditional feature-based systems rely on feature extraction, later using these features 
to train machine learning models for classification. Amongst them, Ling and Weld 
\cite{DBLP:conf/aaai/LingW12} developed FiGER, that uses hand-crafted features to develop 
a multi-label, multi-class perceptron classifier. Yosef et al., \cite{DBLP:conf/acl/YosefBHSW13} 
developed HYENA, i.e., a hierarchical type classification model using hand-crafted features in 
combination with the SVM classifier. Gillick et al., \cite{DBLP:journals/corr/GillickLGKH14} 
proposed context-dependent fine-grained typing using hand-crafted features along with logistic 
regression classifier. Shimaoka et al., \cite{DBLP:conf/akbc/ShimaokaSIR16} developed neural 
architecture for fine-grained entity typing using a combination of automated and hand-crafted 
features.

Embedding models use widely available embedding resources with customized loss functions to form 
classification models. Yogatama et al., \cite{DBLP:conf/acl/YogatamaGL15} used embeddings along 
with Weighted Approximate Rank Pairwise (WARP) loss. Ren et al., \cite{DBLP:conf/emnlp/RenHQHJH16} 
proposed AFET that uses different set of loss functions to model the clean and the noisy entity 
mentions. Abhishek et al., \cite{DBLP:conf/eacl/AbhishekAA17} proposed end-to-end architecture to 
jointly embed the mention and the label embeddings. Xin et al., \cite{DBLP:conf/emnlp/XinZH0S18} 
used language models to compute the compatibility between the context and the entity type prior 
to entity typing. Choi et al., \cite{DBLP:conf/acl/LevyZCC18} proposed ultra-fine entity typing 
encompassing more than 10,000 entity types. They used crowd-sourced data along with the distantly 
supervised data for model training.

~\revise{Especially noteworthy amongst the embedding models are the graph convolution networks, 
introduced in recent past, that extend the concept of convolutions from regular-structured grids to 
graphs~\cite{DBLP:conf/iclr/KipfW17}}. Ali et al., \cite{DBLP:conf/aaai/AliSLW20} 
proposed attentive convolutional network for fine-grained entity typing. 
Nickel et al., \cite{DBLP:conf/nips/NickelK17} illustrated 
the benefits of hyperbolic geometry for embedding the graph structured data. 
Chami et al., \cite{DBLP:conf/nips/ChamiYRL19} combined graph convolutions with the 
hyperbolic geometry. L{\'{o}}pez et al., \cite{DBLP:conf/rep4nlp/LopezHS19} used hyperbolic 
geometry for ultra-fine entity typing. To the best of our knowledge, we are the first to 
explore the combined benefits of the graph convolution networks in relation with the 
hyperbolic geometry for \FGNET{} over distantly supervised noisy data.

%% file: Chapter3.tex
\section{Proposed Approach}

\subsection{Problem Definition}
In this paper, we present a multi-class, multi-label entity typing system using distantly supervised 
data to classify an entity mention into a set of fine-grained entity types. Specifically, we propose 
attentive type-specific contextual aggregation in the hyperbolic space to fine-tune the mention's 
encodings learnt over noisy data prior to entity typing. 
We assume the availability of training corpus $C_{train}$ acquired via distant supervision, and 
manually labeled test corpus $C_{test}$. Each corpus $C$ (train/test) encompasses a set of sentences. 
For each sentence, the contextual token $\{c_i\}_{i=1}^N$, the mention spans $\{m_i\}_{i=1}^N$ 
(corresponding to the entity mentions), and the candidate type labels $\{t_i\}_{i=1}^N \in \{0,1\}^{T}$ 
($T$-dimensional vector with $t_{i,x} = 1$ if $x^{th}$ type corresponds to the true label and 
zero otherwise) have been priorly identified. The type labels are inferred from type hierarchy 
in the knowledge base $\psi$ with the schema $T_{\psi}$. 
Similar to Ren et al.,~\cite{DBLP:conf/emnlp/RenHQHJH16}, we bifurcate the training data $D_{tr}$ 
into clean $D_{tr\text{-}clean}$ and noisy $D_{tr\text{-}noisy}$, if the corresponding mention's 
type-path follows a single path in the type-hierarchy $T_{\psi}$ or otherwise. Following the 
type-path in Figure~\ref{fig:noisy_data} (ii), a mention with labels \emph{\{person, author\}}
 will be considered as clean, whereas, a mention with labels \emph{\{person, president, author\}} 
 will be considered as noisy. 

\subsection{Overview}

\begin{figure*}[t!]
    \centering
    \captionsetup{justification=centering}
    \resizebox{0.93\linewidth}{!}{
        \includegraphics[]{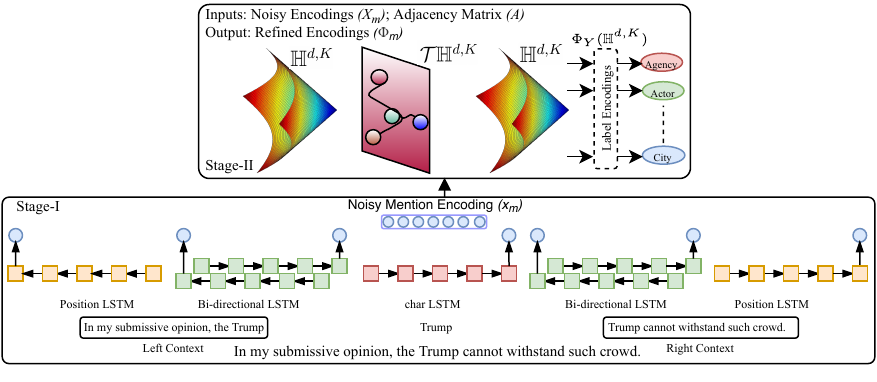}} 
    \caption{Proposed model, i.e., \FGETHR{}, stage-I learns mention's encodings based 
    	on local sentence-specific context, stage-II refines the encodings learnt in 
    	stage-I in the hyperbolic space.}
    \label{fig:proposed_approach}
\end{figure*}

Our proposed model, \FGETHR{}, follows a two-step approach, labeled as stage-I and stage-II in the 
Figure~\ref{fig:proposed_approach}. Stage-I follows text encoding pipeline to generate mention's 
encoding in relation with its context. Stage-II is focused on label noise reduction, for this, 
we map the mention's encoding (from stage-I) in the hyperbolic space and use a graph to share 
aggregated type-specific contextual information along the type-hierarchy in order to refine the 
mention encoding. Finally, the refined mention encoding is embedded along with the label encodings 
in the hyperbolic space for entity typing. Details of each stage are given in the following sub-sections.

\subsection{Stage-I (Noisy Mention Encoding)}
Stage-I follows a standard text processing pipeline using multiple LSTM 
networks~\cite{DBLP:journals/neco/HochreiterS97} to encode the entity mention 
in relation with its context. Individual components of stage-I are explained as follows:

\paragraph{Mention Encoding:} 
We use LSTM network to encode the character sequence corresponding to the mention tokens. 
We use $\phi_{e} = [\largevec{men}] \in \mathbf{R}^{e} $ to represent the encoded mention's tokens.

\paragraph{Context Encoding:}
For context encoding, we use multiple Bi-LSTM networks to encode the tokens corresponding 
to the left and the right context of the entity mention. 
We use $\phi_{c_{l}}$ = $[\cev{c_l}; \largevec{c_{l}}] \in \mathbf{R}^{c}$ and 
$\phi_{c_{r}}$ = $[\cev{c_r}; \largevec{c_{r}}] \in \mathbf{R}^{c}$ to represent the 
encoded left and the right context respectively.

\paragraph{Position Encoding:}
For position encoding, we use LSTM network to encode the position\revise{s} of the 
left and the right contextual tokens. We use $\phi_{p_{l}}$ = $[\cev{l_p}] \in \mathbf{R}^{p}$ 
and ; $\phi_{p_{r}} = [\largevec{r_p}] \in \mathbf{R}^{p}$ to represent the encoded position 
corresponding to the mention's left and the right context.

\paragraph{Mention Encodings:}
Finally, we concatenate all the mention-specific encodings to get L-dimensional context-dependent 
noisy mention encoding: $x_{m} \in \mathbf{R}^{L}$, where $ L = e + 2 * c + 2 * p$.

\begin{equation}
\label{concat_eq}
x_{m} = [\phi_{{p_{l}}} ; \phi_{c_{l}} ; \phi_{e} ; \phi_{c_{r}} ; \phi_{p_{r}}]    
\end{equation}

\subsection{Stage-II (Fine-tuning the Mention Encodings)}

Stage-II is focused on alleviating the label noise. Underlying assumption in combating the 
label noise is that the contextually similar mentions should get similar type labels. For this, 
we form a graph to cluster contextually-similar mentions and employ hyperbolic geometry to 
share the contextual information along the type-hierarchy. As shown in 
Figure~\ref{fig:proposed_approach}, the stage-II follows the following pipeline:

\begin{enumerate}
    \item Construct a graph~\revise{G} such that contextually and semantically similar mentions 
    end-up being the neighbors in the graph.
    
    \item Use exponential map to project the noisy mention encodings from stage-I 
    to the hyperbolic space.
    
    \item In the hyperbolic space, use the corresponding exponential and logarithmic 
    transformations to perform the core operations, i.e., \textit{(i)} linear 
    transformation, and \textit{(ii)} contextual aggregation,  required to fine-tune 
    the encodings learnt in stage-I prior to entity typing.
    
\end{enumerate}

We~\revise{analyze the performance} of \FGETHR{} using two different models in the 
hyperbolic space, i.e., the Hyperboloid $(\mathbb{H}^{d})$ 
and the Poincar\'e-Ball $(\mathbb{D}^{d})$. In the following sub-sections, 
we provide the mathematical formulation for the Hyperboloid model of the hyperbolic 
space. Similar formulation can be designed for the Poincar\'e-Ball model.


\subsubsection{Hyperboloid Model}

$d$-dimensional hyperboloid model of the hyperbolic space (denoted by $\mathbb{H}^{d, K}$) 
is a space of constant negative curvature ${-1}/{K}$, with $\mathcal{T}_{\textbf{p}}\mathbb{H}^{d,K}$ 
as the euclidean tangent space at point $\textbf{p}$, such that:

\begin{align}   
\mathbb{H}^{d,K} &= \{\textbf{p} \in \mathbb{R}^{d+1}:\langle\textbf{p},\textbf{p}\rangle = -K, p_{0} > 0 \} \nonumber \\   
\mathcal{T}_{\textbf{p}}\mathbb{H}^{d,K} &= {\textbf{r} \in \mathbb{R}^{d+1} : \langle \textbf{r},\textbf{p}\rangle_{\mathcal{L}}=0}
\end{align}

where $\langle,.,\rangle_{\mathcal{L}}: \mathbb{R}^{d+1} \times \mathbb{R}^{d+1} \rightarrow \mathbb{R}$ 
denotes the Minkowski inner product, with $\langle\textbf{p},\textbf{q}\rangle_{\mathcal{L}} = -p_{0}q_{0} + p_{1}q_{1} + ... + p_{d}q_{d}$.

\paragraph{Geodesics and Distances:}
For two points $\textbf{p}$, $\textbf{q} \in \mathbb{H}^{d,K}$, the distance function between them is given by:

\begin{align}
d_{\mathcal{L}}^{K}(\textbf{p},\textbf{q}) = \sqrt{K} \text{arccosh}(-\langle \textbf{p},\textbf{q}\rangle_{\mathcal{L}}/K)
\end{align}

\paragraph{Exponential and Logarithmic maps:}
We use exponential and logarithmic maps for mapping to and from the hyperbolic and the tangent space 
respectively. Formally, given a point $\textbf{p} \in \mathbb{H}^{d,K}$ and tangent vector 
$\textbf{t} \in \mathcal{T}_{\textbf{p}}\mathbb{H}^{d,K}$, the exponential map 
$\exp_{\textbf{p}}^{K}:\mathcal{T}_{\textbf{p}}\mathbb{H}^{d,K}\rightarrow \mathbb{H}^{d,K}$ 
assigns a point to $\textbf{t}$ such that $\exp_{\textbf{p}}^{K}(\textbf{t}) = \gamma(1)$, where 
$\gamma$ is the geodesic curve that satisfies  $\gamma(0) = \textbf{p}$ and $\dot{\gamma} = \textbf{t}$. 

The logarithmic map $(\log^{K}_{\textbf{p}})$ being the bijective inverse  maps a point in 
hyperbolic space to the tangent space at $\textbf{p}$. We use the following equations for the 
exponential and the logarithmic maps:

\begin{equation}
\label{Eq:hyperboloid_exp}
\exp_{\textbf{p}}^{K}(\textbf{v}) = \cosh(\frac{||\textbf{v}||_{\mathcal{L}}}{\sqrt{K}})\textbf{p} + \sqrt{K}\sinh(\frac{||\textbf{v}||_{\mathcal{L}}}{\sqrt{K}})\frac{\textbf{v}}{||\textbf{v}||_{\mathcal{L}}}
\end{equation}

\begin{equation}
\label{Eq:hyperboloid_log}
\log^{K}_{\textbf{p}}(\textbf{q}) = d_{\mathcal{L}}^{K}(\textbf{p},\textbf{q}) \frac{\textbf{q}+\frac{1}{K}<\textbf{p},\textbf{q}>_{\mathcal{L}}\textbf{p}} {||\textbf{q}+\frac{1}{K}<\textbf{p},\textbf{q}>_{\mathcal{L}}\textbf{p}||_{\mathcal{L}}}
\end{equation}

\subsubsection{Graph Construction}

The end-goal of graph construction is to group the entity mentions in such a way that contextually similar 
mentions~\revise{end up being neighbours in the graph by forming edges}. Given the fact, the euclidean 
embeddings are better at capturing the semantic aspects of the text data~\cite{DBLP:conf/textgraphs/DhingraSNDD18}, 
we opt to use deep contextualized embeddings in the euclidean domain~\cite{DBLP:conf/naacl/PetersNIGCLZ18} 
for the graph construction. For each entity type, we average out corresponding $1024d$ embeddings for all 
the mentions in the training corpus $C_{train}$, to learn prototype vectors for each entity type, 
i.e., $\{prototype_{t}\}_{t=1}^{T}$. Later, for each entity type $t$, we capture type-specific 
confident~\revise{entity} mention candidates $cand_{t}$, following the criterion: 
$\{cand_{t} = cand_{t} \cup men \text{ if } (cos(men, \{Prototype_{t}\}) \geq \delta)$  
$ \forall men \in C;  \forall{} t \in T \}$, where $\delta$ is a threshold.
Finally, we form pairwise edges for all the mention candidates corresponding to 
each entity-type, i.e., $\{cand\}_{t=1}^{T}$, to construct the graph $G$, with adjacency matrix $A$.
\revise{
Formulating the graph in this particular manner allows similar mentions (i.e., sharing 
similar context) to be clustered around each other by forming edges in the graph, 
which facilitates the information passing across the noisy entity mentions. The 
granularity of the information shared may be controlled by edge weights.
}

\subsubsection{Mapping Noisy Mention Encodings to the Hyperbolic space}

The mention encodings learnt in the stage-I are noisy, as they are learnt over distantly supervised 
data. These encodings lie in the euclidean space, and in order to refine them, we first map them 
to the hyperbolic space, where we may best exploit the fine-grained type hierarchy in relation 
with the type-specific contextual clues~\revise{(using $G$)} to fine-tune these encodings as an 
aggregate of contextually-similar neighbours.

Formally, let $\mathbf{p}^{E} = X_{m} \in \mathbf{R}^{N \times L}$ be the matrix corresponding to the 
noisy mentions' encodings in the euclidean domain. We consider $ o = \{\sqrt{K}, 0,...,0\}$ as a 
reference point (origin) in a d-dimensional Hyperboloid with curvature 
$-1/K \, (\mathbb{H}^{d,K})$; $(0,\textbf{p}^{E})$ as a point in the tangent space 
$(\mathcal{T}\mathbb{H}^{d,K})$, and map it to $\textbf{p}^{H} \in \mathbb{H}^{d,K}$ using the 
exponential map given in Equation~\eqref{Eq:hyperboloid_exp}, as follows:

\begin{align}
\textbf{p}^{H} &= \exp^{K}((0,\textbf{p}^{E})) \nonumber \\
\exp^{K}((0,\textbf{p}^{E})) &= \Big( \sqrt{K} \cosh\Big(\frac{||\textbf{p}^{E}||_{2}}{\sqrt{K}} \Big), \nonumber \\ 
&  \sqrt{K}\sinh\Big(\frac{||\textbf{p}^{E}||_{2}}{\sqrt{K}}\Big)\frac{\textbf{p}^{E}}{||\textbf{p}^{E}||_{2}}\Big)
\end{align}


\subsubsection{Linear Transformation}
In order to perform linear transformation operation on the noisy mention encodings, i.e., \textit{(i)} 
multiplication by weight matrix $\textbf{W}$, and \textit{(ii)} addition of bias vector $\textbf{b}$, 
we rely on the exponential and the logarithmic maps. For multiplication with the weight matrix, firstly, 
we apply logarithmic map on the encodings in the hyperbolic space, i.e., $\textbf{p}^{H} \in \mathbb{H}^{d,K}$, 
in order to project them to $\mathcal{T}\mathbb{H}^{d,K}$.
This projection is then multiplied by the weight matrix $W$, and the resultant vectors are projected 
back to the manifold using the exponential map. For a manifold with curvature constant $K$, these 
operations can be summarized in the equation, given below:
\begin{equation}
W \otimes \textbf{p}^{H} = \exp^{K}(W\log^{K}(\textbf{p}^{H})) 
\end{equation}

For bias addition, we rely on parallel transport, let $\textbf{b}$ be the bias vector in 
$\mathcal{T}\mathbb{H}^{d,K}$, we parallel transport $\textbf{b}$ along the tangent space and 
finally map it to the manifold. Formally, let $\textbf{T}^{K}_{\textbf{o} \rightarrow \textbf{p}^{H}}$ 
represent the parallel transport of a vector from $\mathcal{T}_{\textbf{o}}\mathbb{H}^{d,K}$ to 
$\mathcal{T}_{\textbf{x}^{H}}\mathbb{H}^{d,K}$, we use the following equation for the bias addition:

\begin{equation}
\textbf{p}^{H} \oplus \textbf{b} = \exp^{K}_{\textbf{x}^{H}}(\textbf{T}^{K}_{o \rightarrow \textbf{p}^{H}}(\textbf{b})) 
\end{equation}

\subsubsection{Type-Specific Contextual Aggregation}
Aggregation is a crucial step for noise reduction in \FGNET{}, it helps to smoothen the type-label 
by refining/fine-tuning the noisy mention encodings by accumulating information from contextually 
similar neighbours lying at multiple hops. Given the graph $G$, with nodes $(V)$ being the entity 
mentions, we use the pairwise embedding vectors along the edges of the graph to compute the 
attention weights $\eta_{ij} = cos(men^{i}, men^{j}) \forall (i,j) \in V$.
In order to perform the aggregation operation, we first use the logarithmic map to project the 
results of the linear transformation from hyperbolic space to the tangent space.
Later, we use the neighbouring information contained in $G$ to compute the refined mention 
encoding as attentive aggregate of the neighbouring mentions. 
Finally, we map these results back to the manifold using the exponential map $\exp^{K}$. 
Our methodology for contextual aggregation is summarized in the following equation:

\begin{equation}
AGG_{cxtx} (\textbf{p}^{H})_{i} = \exp^{K}_{\textbf{x}^{H}_{i}} \Big( \sum_{j \in \textit{N}(i)} (\widetilde{\eta_{ij} \odot A}) \log^{K}(\textbf{p}^{H}_{j}) \Big)
\end{equation}

where $\widetilde{\eta_{ij} \odot A}$ is the Hadamard product of the attention weights and the 
adjacency matrix $A$. It accommodates the degree of contextual similarity among the mention pairs in $G$.

\subsubsection{Non-Linear Activation}
Contextually aggregated mention's encoding is finally passed through a non-linear activation function 
$\sigma$ ($\Relu$ in our case). For this, we follow similar steps, i.e., \textit{(i)} map the 
encodings to the tangent space,
\textit{(ii)} apply the activation function in the tangent space, 
\textit{(iii)} map the results back to the hyperbolic space using exponential map. These steps are 
summarized in the following equation:

\begin{equation}
\sigma(\textbf{p}^{H}) = \exp^{K}(\sigma(\log^{K}(\textbf{p}^{H})))
\end{equation}
\subsection{Complete Model}
We combine the above-mentioned steps to get the refined mention encodings at \textit{lth}-layer 
$\textbf{z}_{out}^{l,H}$ as follows:

\begin{align}
\textbf{p}^{l,H} &= W^{l} \otimes \textbf{p}^{l-1,H} \oplus \textbf{b}^{l}\text{;} \nonumber \\ 
\textbf{y}^{l,H} &= AGG_{cxtx}(\textbf{p}^{l,H})\text{;} \,\,\, 
\textbf{z}^{l,H}_{out} = \sigma(\textbf{y}^{l,H})
\end{align}

Let $\textbf{z}_{out}^{l,H} \in \mathbb{H}^{d,K}$ correspond to the refined mentions' encodings
 hierarchically organized in the hyperbolic space. We embed them along with the fine-grained type 
 label encodings $\{\phi_{t}\}_{t=1}^{T} \in \mathbb{H}^{d}$. For that we learn a function 
 $f(\textbf{z}^{l,H}_{out},\phi_{t}) = \phi_{t}^{T}\times\textbf{z}^{l,H} + bias_{t}$, and 
 separately learn the loss functions for the clean and the noisy mentions. 

\paragraph{Loss for clean mentions:} In order to model the clean entity mentions $D_{tr\text{-}clean}$, 
we use a margin-based loss to embed the refined mention encodings close to the true type labels 
($T_y$), and push it away from the false type labels ($T_{y^{'}}$). The loss function is summarized as follows:

\begin{align}
L_{clean} &= \sum_{t \in T_y} \Relu(1-f(\textbf{z}^{l,H}_{out},\phi_{t})) + \nonumber \\
& \sum_{t^{'} \in T_{y^{'}}} \Relu(1 + f(\textbf{z}^{l,H}_{out},\phi_{t^{'}}))
\end{align}

\paragraph{Loss for noisy mentions:} In order to model the noisy entity mentions $D_{tr\text{-}noisy}$, 
we use a variant of above-mentioned loss function to embed the mention close to most relevant type 
label $t^{*}$, where $t^{*} = \argmax_{t \in T_y}f(\textbf{z}^{l,H}_{out},\phi_{t})$, among the 
set of noisy type labels $(T_y)$ and push it away from the irrelevant type labels ($T_{y^{'}}$). 
The loss function is mentioned as follows:

\begin{align}
L_{noisy} &= \Relu(1-f(\textbf{z}^{l,H}_{out},\phi_{t^{*}})) + \nonumber \\
& \sum_{t^{'} \in T_{y^{'}}} \Relu(1 + f(\textbf{z}^{l,H}_{out},\phi_{t^{'}})) 
\end{align}
Finally, we minimize $L_{clean} + L_{noisy}$ as the final loss function of the \FGETHR{}.

%% file: Chapter4.tex
\section{Experimentation}

\subsection{Dataset}
We evaluate our model using a set of publicly available datasets for \FGNET{}. We 
chose these datasets because they contain fairly large proportion of test instances 
and corresponding evaluation will be more concrete. Statistics of these dataset is 
shown in Table~\ref{tab:dataset_FG_GCN}. These datasets are explained as follows:

\paragraph{BBN:} Its training corpus is acquired from the Wall Street Journal annotated 
by~\cite{weischedel2005bbn} using DBpedia Spotlight.

\paragraph{OntoNotes:} It is acquired from newswire documents contained in the 
OntoNotes corpus~\cite{weischedel2011ontonotes}. The training data is mapped to Freebase 
types via DBpedia Spotlight~\cite{DBLP:conf/i-semantics/DaiberJHM13}. The testing data 
is manually annotated by Gillick et al.,~\cite{DBLP:journals/corr/GillickLGKH14}.

\begin{table}[t!]
    \centering
    \resizebox{0.85\linewidth}{!}{
        \begin{tabular}{l | rr}
            \hline
            Dataset & BBN  & OntoNotes \\
            \hline
            Training Mentions & 86078  & 220398 \\
            Testing Mentions & 13187   & 9603 \\
            \% clean mentions (training) & 75.92    & 72.61 \\
            \% clean mentions (testing) & 100  & 94.0 \\
            Entity Types & 47  & 89 \\
            \hline
        \end{tabular}}
        \caption{Fine-Grained Named Entity Typing data sets}
        \label{tab:dataset_FG_GCN}
    \end{table}

\subsection{Experimental Settings}
In order to set up a fair platform for comparative evaluation, we use the same data settings 
(training, dev and test splits) as used by all the models considered as baselines in 
Table~\ref{tab:Results}. All the experiments are performed using Intel Gold 6240 CPU with 
256 GB main memory.

\paragraph{Model Parameters:}
For stage-I, the hidden layer size of the context and the position encoders is set to 100d. 
The hidden layer size of the mention character encoder is 200d. Character, position and 
label embeddings are randomly initialized. We report the model performance using 300d 
Glove~\cite{DBLP:conf/emnlp/PenningtonSM14} and 1024d deep 
contextualized embeddings~\cite{DBLP:conf/naacl/PetersNIGCLZ18}.
 
For stage-II, we construct graphs with 5.4M ( using $\delta = 0.75$) and 0.6M (using $\delta= 0.70$) 
edges for BBN and OntoNotes respectively. Curvature constant of the hyperbolic space is set 
to $K=1$. All the models are trained using Adam optimizer~\cite{DBLP:journals/corr/KingmaB14} 
with learning rate = 0.001.

\subsection{Model Comparison}

We evaluate \FGETHR{} against the following baseline models: (i) Figer~\cite{DBLP:conf/aaai/LingW12}; 
(ii) Hyena~\cite{DBLP:conf/acl/YosefBHSW13}; (iii) AFET, AFET-NoCo and AFET-NoPa~\cite{DBLP:conf/emnlp/RenHQHJH16}; (iv) Attentive~\cite{DBLP:conf/akbc/ShimaokaSIR16}; (v) FNET~\cite{DBLP:conf/eacl/AbhishekAA17}; 
(vi) NFGEC + LME~\cite{DBLP:conf/emnlp/XinZH0S18}; and (vii) FGET-RR~\cite{DBLP:conf/aaai/AliSLW20}. 
For performance comparison, we use the scores reported in the original papers, as they are computed 
using a similar data settings as that of ours.

Note that we do not compare our model against~\cite{DBLP:conf/acl/LevyZCC18, DBLP:conf/rep4nlp/LopezHS19} 
because these models use crowd-sourced data in addition to the distantly supervised data for model 
training. Likewise, we exclude \cite{DBLP:conf/naacl/XuB18} from evaluation because Xu and Barbosa 
changed the fine-grained problem definition from multi-label to single-label
classification problem. This makes their problem settings different from that of ours and the end 
results are no longer comparable.

\subsection{Main Results}

The results of the proposed model are shown in Table~\ref{tab:Results}. For each data set, we boldface 
the best scores with the existing state-of-the art underlined. These results show that \FGETHR{} 
outperforms the existing state-of-the-art models by a significant margin. For the BBN data,~\FGETHR{} 
achieves 3.5\%, 1.2\% and 1.5\% improvement in strict accuracy, mac-F1 and mic-F1 respectively, 
\revise{compared to the previous best, i.e., FGET-RR}. For OntoNotes,~\FGETHR{} 
improves the mac-F1 and mic-F1 scores by 1.2\% and 1.6\%.

These results show that \FGETHR{} offers multi-faceted benefits, i.e., using hyperbolic space in 
combination with the graphs to encode the fine-grained type hierarchy, while at the same time 
catering to noise in the best possible way.
\revise{
	This setting is best suited for~\FGNET{} over distantly supervised data, especially because it 
	allows \FGETHR{} to perform augmented context sharing along the type hierarchy which plays a vital 
	role for label smoothing at different levels of granularity.
}

\begin{table*}[t!]
    \centering
    \resizebox{.75\linewidth}{!}{%
        \begin{tabular}{l|lll|lll}
            \hline
            & \multicolumn{3}{c}{OntoNotes} & \multicolumn{3}{c}{BBN}  \\
            \hline
            \hline
            & strict   & mac-F1   & mic-F1  & strict & mac-F1 & mic-F1 \\
            \hline
            \hline
            \textbf{FIGER}~\cite{DBLP:conf/aaai/LingW12}     & 0.369    & 0.578    & 0.516   & 0.467  & 0.672  & 0.612  \\
            \textbf{HYENA}~\cite{DBLP:conf/acl/YosefBHSW13}    & 0.249    & 0.497    & 0.446   & 0.523  & 0.576  & 0.587  \\
            \textbf{AFET-NoCo}~\cite{DBLP:conf/emnlp/RenHQHJH16}   & 0.486    & 0.652    & 0.594   & 0.655  & 0.711  & 0.716  \\
            \textbf{AFET-NoPa}~\cite{DBLP:conf/emnlp/RenHQHJH16}   & 0.463    & 0.637    & 0.591   & 0.669  & 0.715  & 0.724  \\
            \textbf{AFET-CoH}~\cite{DBLP:conf/emnlp/RenHQHJH16}   & 0.521    & 0.680     & 0.609   & 0.657  & 0.703  & 0.712  \\
            \textbf{AFET}~\cite{DBLP:conf/emnlp/RenHQHJH16}    & {0.551}    & 0.711    & 0.647   & {0.670}  & 0.727  & 0.735  \\
            \textbf{Attentive}~\cite{DBLP:conf/akbc/ShimaokaSIR16}   & 0.473    & 0.655    & 0.586   & 0.484  & 0.732  & 0.724  \\
            \textbf{FNET-AllC}~\cite{DBLP:conf/eacl/AbhishekAA17}   & 0.514    & 0.672    & 0.626   & 0.655  & 0.736  & 0.752  \\
            \textbf{FNET-NoM}~\cite{DBLP:conf/eacl/AbhishekAA17}    & 0.521    & 0.683    & 0.626   & 0.615  & 0.742  & 0.755  \\
            \textbf{FNET}~\cite{DBLP:conf/eacl/AbhishekAA17}   & 0.522    & 0.685   & 0.633  & 0.604   & 0.741  & 0.757  \\
            \textbf{NFGEC+LME}~\cite{DBLP:conf/emnlp/XinZH0S18}  & 0.529    & {0.724}   & {0.652}  & 0.607     &  {0.743}   & {0.760}  \\
            \textbf{FGET-RR}\cite{DBLP:conf/aaai/AliSLW20} (Glove)  &  {0.567}  &  {0.737}  &  {0.680}     & \underline{0.740}  & {0.811}  & {0.817}\\
            \textbf{FGET-RR}\cite{DBLP:conf/aaai/AliSLW20} (ELMO)  &  \underline{0.577}  &  \underline{0.743}  & \underline{0.685}  & 0.703  & \underline{0.819}  & \underline{0.823} \\
            \hline
            \hline
            \textbf{\FGETHR{}} (Hyperboloid + Glove)    &  \textbf{0.580}  &  0.738  &  0.685     & \textbf{0.766}  & 0.828  & \textbf{0.835} \\
            \textbf{\FGETHR{}} (Hyperboloid + ELMO)     &  0.575  &  \textbf{0.752}  &  \textbf{0.696}     & 0.712  & 0.824  & 0.823 \\
            \textbf{\FGETHR{}} (Poincar\'e-Ball + Glove)  &  0.579  &  0.741  &  0.684     & 0.760  & \textbf{0.829}  & 0.833 \\
            \textbf{\FGETHR{}} (Poincar\'e-Ball + ELMO)   &  0.573  &  0.740  &  0.685  & 0.698  & 0.828  & 0.830 \\
            \hline
        \end{tabular}}
        \caption{\FGNET{} performance comparison against baseline models}
        \label{tab:Results}
    \end{table*}

\subsection{Ablation Study}
\label{sec:abl_study}
\revise{
In the following sub-sections, we perform in-depth analysis of \FGETHR{}, including: 
(i) Role of adjacency graph $(G)$; 
(ii) Effectiveness of hyperbolic geometry;
(iii) Impact of stage-II;
(iv) Analysis of label vectors; and 
(v) Error cases.}

\subsubsection{Role of adjacency graph $(G)$}
We analyze the performance of \FGETHR{} using variants of the adjacency graph, 
including: \textit{(i)} randomly generated adjacency graph of approximately the same size as 
$G$: $\FGETHR{} \, (R)$, \textit{(ii)} unweighted adjacency graph: $\FGETHR{} $ $\, (A)$, and 
\textit{(iii)} pairwise contextual similarity as the attention weights $\FGETHR{} \,(\widetilde{\eta \odot A})$. 
The results in Table~\ref{tab:Ablation_study} show that for the given model architecture, 
the performance improvement (correspondingly noise-reduction) can be attributed to using 
the appropriate adjacency graph. 

A drastic reduction in the model performance for $\FGETHR{} \, (R)$ shows that once the 
contextual similarity structure of the~\revise{adjacency graph is lost, the label-smoothing 
is no longer effective to combat the label-noise.
This is also evident from a relatively higher performance by the models: $\FGETHR{} \, (A)$, 
and $\FGETHR{}$ $\, (\widetilde{\eta \odot A})$ using unweighted adjacency graph $(A)$ and 
attention weights $(\widetilde{\eta \odot A})$ respectively. 

Especially noteworthy is the impact of the attention weights $(\widetilde{\eta \odot A})$,
which strongly indicates that, for label de-noising within each type-specific contextual 
cluster, each mention has a different impact on its neighbouring mentions in $G$ 
depending upon the degree of their contextual similarities. It, moreover, confirms that 
$\FGETHR{} \, (\widetilde{\eta \odot A})$ indeed incorporates the required type-specific 
contextual clusters at the needed level of granularity to effectively smoothen the 
noisy labels prior to the entity typing.

}

\begin{table}[b!]
    \centering
    \captionsetup{justification=centering}
    \resizebox{.98\linewidth}{!}{%
    \begin{tabular}{l|lll|lll}
        \hline
        \multicolumn{1}{c}{\multirow{3}{*}{Model}} & \multicolumn{3}{c}{OntoNotes} & \multicolumn{3}{c}{BBN}        \\
        \hline
        \multicolumn{1}{c|}{}                       & strict  & mac-F1   & mic-F1   & strict   & mac-F1   & mic-F1   \\
        \hline
        
        \hline
        $\FGETHR{} \, (R)$                           & 0.484   &  0.643   & 0.597    & 0.486  & 0.647 & 0.653        \\
        $\FGETHR{} \, (A)$                           & 0.531   &  0.699   & 0.632    & 0.735  & 0.808 & 0.815        \\
        $\FGETHR{} \, (\widetilde{\eta \odot A})$    & 0.580   & 0.738    & 0.685    & 0.766  & 0.828 & 0.835        \\
        \hline
        \multicolumn{1}{c}{}                       & \multicolumn{6}{c}{Hyperboloid ($\mathbb{H}^{d}$)}                                    \\
        \hline
        $\FGETHR{} \, (R)$                         & 0.490   & 0.665    & 0.608    & 0.633    & 0.704    & 0.724    \\
        $\FGETHR{} \, (A)$                         & 0.571   & 0.737    & 0.679    & 0.746    & 0.814    & 0.822    \\
        $\FGETHR{} \, (\widetilde{\eta \odot A})$  & 0.579   & 0.741    & 0.684    & 0.760    & 0.829    & 0.833    \\
        \hline
        \multicolumn{1}{c}{}                       & \multicolumn{6}{c}{Poincar\'e-Ball ($\mathbb{D}^{d}$)}                                  \\
        \hline
    \end{tabular}}
    \caption{\FGETHR{} performance comparison using different adjacency matrices and Glove Embeddings}
    \label{tab:Ablation_study}
\end{table}

\subsubsection{Effectiveness of hyperbolic geometry}
\label{subec:effectiveness}
In order to verify the effectiveness of refining the mention encodings in the hyperbolic space 
(stage-II), we perform label-wise performance analysis for the dominant labels in the BBN dataset. 
Corresponding results for the Hyperboloid and the Poincar\'e-Ball model (in Table~\ref{tab:label_wise}) 
show that \FGETHR{} outperforms the existing state-of-the-art, i.e., FGET-RR by Ali et al.,~\cite{DBLP:conf/aaai/AliSLW20}, 
achieving higher F1-scores across all the labels. Note that \FGETHR{} can achieve higher 
performance for the base type labels: \{e.g., \emph{``/Person", ``/Organization", ``/GPE"} etc.,\}, 
as well as other type labels down in the hierarchy, \{e.g., \emph{``/Organization/Corporation", 
	``/GPE/City"} etc.,\}. For \{\emph{``Organization"} and \emph{``Corporation"}\} \FGETHR{} 
achieves a higher F1=0.896 and F1=0.855 respectively, compared to the F1=0.881 and F1=0.844 by 
FGET-RR. This is made possible because embedding in the hyperbolic space enables type-specific 
context sharing at each level of the type hierarchy by appropriately adjusting the norm of the label vector.

\begin{table}[h!]
        \centering
        \captionsetup{justification=centering}
        \resizebox{1.0\linewidth}{!}{%
    \begin{tabular}{l|c|ccc|ccc|ccc}
        \hline
        \multirow{2}{*}{Labels} & \multirow{2}{*}{Support} & \multicolumn{3}{c|}{FGET-RR~\cite{DBLP:conf/aaai/AliSLW20}} & \multicolumn{3}{c}{\FGETHR{} (Poincar\'e-Ball)} & \multicolumn{3}{|c}{\FGETHR{} (Hyperboloid)} \\
         &      & Prec   & Rec         & F1          & Prec          & Rec          & F1           & Prec         & Rec          & F1           \\
        \hline
        /Organization        &   45.30\%     & 0.924        & 0.842       & 0.881       & 0.916         & 0.876        & \textbf{0.896}        & 0.926        & 0.860        & 0.891        \\
        /Org/Corporation  &   35.70\%     & 0.921        & 0.779       & 0.844       & 0.903         & 0.812        & \textbf{0.855}        & 0.908        & 0.801        & 0.851        \\
        /Person                    &   22.00\%     & 0.86         & 0.886       & 0.872       & 0.876         & 0.902        & \textbf{0.889}        & 0.843        & 0.911        & 0.876        \\
        /GPE                       &   21.30\%     & 0.924        & 0.845       & 0.883       & 0.92          & 0.868        & 0.893        & 0.924        & 0.885        & \textbf{0.904}        \\
        /GPE/City                  &   9.17\%      & 0.802        & 0.767       & 0.784       & 0.806         & 0.750        & 0.777                 & 0.804        & 0.795        & \textbf{0.799}        \\
        \hline    
    \end{tabular}}
    \caption{Label-wise Precision, Recall and F1 scores for the BBN data compared with FGET-RR~\cite{DBLP:conf/aaai/AliSLW20}}
    \label{tab:label_wise}
\end{table}

To further strengthen our claims regarding the effectiveness of using hyperbolic space for \FGNET{}, 
we analyzed the context of the entity types along the type-hierarchy. We observed, for the fine-grained 
type labels, the context is additive and may be arranged in a hierarchical structure with the generic 
terms lying at the root and the specific terms lying along the children nodes. For example, 
\emph{``Government Organization"} being a subtype of \emph{``Organization"} adds tokens similar to 
\{bill, treasury, deficit, fiscal, senate etc., \} to the context of \emph{``Organization"}. 
Likewise, \emph{``Hospital"} adds tokens similar to \{family, patient, kidney, stone, infection 
etc., \} to the context of \emph{``Organization"}.

\subsubsection{Impact of stage-II} We also analyzed the entity mentions corrected especially by the label-smoothing 
process, i.e., the stage-II of \FGETHR{}. For this, we examined the model performance with and 
without the label-smoothing, i.e.,~\revise{we perform entity typing solely based on the noisy 
mention encodings learnt in stage-I.}

For the BBN data, the stage-II~\revise{corrects approximately}~18\% of the 
mis-classifications made by stage-I. For example in the sentence: \emph{``CNW Corp. said the 
	final step in the acquisition of the company has been completed with the merger of \textbf{CNW} 
	with a subsidiary of Chicago \& amp."}, the bold-faced entity mention \textbf{CNW} is labeled 
\{\emph{``/GPE"}\} by stage-I. However, after label-smoothing in stage-II, the label predicted 
by \FGETHR{} is \{\emph{``/Organization/Corporation"}\}, which indeed is the correct label. A 
similar trend was observed for the OntoNotes data set.

This analysis concludes that the \FGETHR{} using a blend of the contextual graphs and the 
hyperbolic space incorporates the right geometry to embed the noisy \FGNET{} data with 
lowest possible distortion. Compared to the euclidean space, the hyperbolic space being a 
non-euclidean space allows the graph volume (number of nodes within a fixed radius) to grow 
exponentially along the hierarchy,~\revise{which} enables the \FGETHR{} to perform label-smoothing by 
forming type-specific contextual clusters across noisy mentions along the type hierarchy.

\subsubsection{Analysis of label vectors}
\label{distant_analysis}
\revise{In order to verify our claims that the hyperbolic space is an optimal 
	choice for fine-grained entity typing with highly correlated entity types
, we analyse the distance among the neighbouring label vectors 
to explore the orientation of these label vectors in the hyperbolic space.} 

We report the nearest neighbours w.r.t the hyperbolic distance for the labels: 
\{\emph{``/Location"} and \emph{``/Organization"}\} 
in the hyperboloid model of the hyperbolic space in Table~\ref{tab:label_dist_FGETHR}. 
The nearest neighbours of the label \{\emph{``/Location"}\} include labels~\revise{hierarchically} 
derived from the base type label: \{\emph{``/Location/River", ``/Location/Lake\_Sea\_Ocean"}\}, and 
labels semantically related to the base type label: \{\emph{``/GPE/State\_Province"}\}.
\revise{
Likewise, the nearest neighbours for the label \{``/Organization"\} also encompasses 
a blend of derived labels: 
\{\emph{``/Organization/Hospital", ``/Organization/Hotel"}\} and related labels:
\{\emph{``/GPE/State\_Province"}\}.

This illustrates that within the hyperbolic space, semantically related and 
hierarchically-organized label vectors are oriented in one particular direction 
away from other irrelevant type labels. 
At the same time exponential growth of the volume in hyperbolic space, as we 
move along the radius, makes it more favourable to place these hierarchically 
organized type labels along a hierarchy, thus allowing customized context 
sharing for each entity type at a much finer level of granularity.}

These findings also correlate with the norm of the label vectors, shown in 
Table~\ref{tab:label_norm} for the Poincar\'e-Ball model. The vector norm of 
the entity types deep in the hierarchy 
\{e.g., \emph{``/Facility/Building", ``/Facility/Bridge", ``/Facility/Highway"} etc., \} is 
greater than that of the base entity type \{ \emph{``/Facility"} \}. A similar trend is 
observed for the fine-grained types: \{\emph{``/Organization/Government", ``/Organization/Political"} etc.,\} 
compared to the base type: \{\emph{``/Organization"}\}. It justifies that \FGETHR{}  
adjusts the norm of the label vector according to the depth of the type-label in the 
label-hierarchy, which allows the model to consequently cluster the type-specific context 
along the hierarchy in an augmented fashion.


\begin{table}[t]
	\centering
    \captionsetup{justification=centering}
	\resizebox{.98\linewidth}{!}{%
		\begin{tabular}{l|c|l|c}
			\hline
			\multicolumn{1}{l|}{Label (/Location)} & Distance  & \multicolumn{1}{l|}{Label (/Organization)} & Distance  \\
			\hline
			/Location 						& 0.0  	 & /Organization             & 0.0  \\
			/Location/River                	& 0.120  & /Organization/Hospital    & 1.362 \\
			/Location/Lake\_Sea\_Ocean      & 0.292  & /Organization/Hotel       & 1.643 \\
			/GPE/State\_Province           	& 0.665  & /GPE/State\_Province      & 1.760 \\
			\hline
	\end{tabular}}
	\caption{\FGETHR{} distance from nearest neighbouring label vectors in the Hyperboloid model of the hyperbolic space ($\mathbb{H}^d$)}
	\label{tab:label_dist_FGETHR}
\end{table}

\subsubsection{Error Cases}
\revise{Finally, }we analyzed the prediction errors of \FGETHR{} and attribute them to the 
following factors:
 
\paragraph{Inadequate Context:} For these~\revise{error} cases, type-labels are dictated entirely 
by the mention tokens, with very little information contained in the context. For example, in the 
sentence: \emph{``The \textbf{IRS} recently won part of its long-running battle against John."}, 
the entity mention \emph{``\textbf{IRS}"} is labeled as \{\emph{``/Organization/Corporation"}\} 
irrespective of any information contained in the mention's context. Limited information 
contained in the mention's context in turn limits the end-performance of \FGETHR{} in 
predicting all possible fine-grained labels thus effecting the recall.
~\revise{We observed, for the BBN data set, roughly 30\% of the errors were caused by the 
	inadequate mention's context.}

\paragraph{Correlated Context:}
\revise{A particular problem associated with the \FGNET{} is the lack of pre-defined 
	set of type labels. For each data set, the fine-grained type hierarchy 
	encompass a blend of semantically correlated type labels with convoluted/in-distinguishable 
	context, also observed in section~\ref{distant_analysis}.

For the BBN data set, we observed: \{\emph{``Actor"} vs \emph{``Artist"}\}; 
\{\emph{``Actor"} vs \emph{``Director"}\}; \{ \emph{``Organization"} vs \emph{``Corporation"}\}; 
\{\emph{``Ship"} vs \emph{``Spacecraft"}\}; \{\emph{``Coach"} vs \emph{``Athlete"}\} etc., 
as some of the correlated entity types with highly convoluted context. 
For example, the context of the entity types \{\emph{``Actor"}\} and \{\emph{``Artist"}\} 
is extremely overlapping, as some of the semantically-related tokens like: 
\{direct, dialogue, dance, acting, etc.,\} appear in the context of each of these 
entity types. Such excessive contextual overlap makes it hard for the \FGETHR{} to delineate 
the decision boundary across these correlated entity types. It leads to false predictions by the 
model thus effecting the precision. For the BBN data set, more than 35\% errors may be attributed 
to the correlated context.
}

\paragraph{Label Bias:} 
~\revise{
	Label bias originating from the training data automatically acquired via distant supervision 
	may result in the label-smoothing (stage-II of \FGETHR{}) to be in-effective. This occurs, 
	specifically, if all the labels originating from the distant supervision are incorrect. 
	For the BBN data approximately 5\% errors may be attributed to the label bias.
}

The rest of the errors may be attributed to the inability of the \FGETHR{} to explicitly deal with 
different word senses, in-depth syntactic analysis, in-adequacy of underlying embedding models 
to handle semantics, etc. We plan to accommodate these aspects in the future work.

\begin{table}[]
	\centering
    \captionsetup{justification=centering}
	\resizebox{.98\linewidth}{!}{%
		\begin{tabular}{l|c|l|c}
			\hline
			\multicolumn{1}{c|}{Label} & Norm  & \multicolumn{1}{c|}{Label} & Norm  \\
			\hline
			/Organization                		& 0.855  & /Facility                 & 0.643 \\
			/Organization/Religious       		& 0.860  & /Facility/Building        & 0.725 \\
			/Organization/Government           	& 0.870  & /Facility/Bridge          & 0.745 \\
			/Organization/Political           	& 0.875  & /Facility/Highway         & 0.815 \\
			\hline
	\end{tabular}}
	\caption{\FGETHR{} Label-norms for the Poincar\'e-Ball model, the norm for the base type-labels 
		is lower than the type-labels deep in the hierarchy}
	\label{tab:label_norm}
\end{table}

\section{Conclusions}
In this paper, we introduced \FGETHR{}, a novel approach that combines the benefits of graph 
structures and hyperbolic geometry to perform entity typing in a robust fashion. 
\FGETHR{} initially learns noisy mention encodings using LSTM networks and constructs a graph 
to cluster contextually similar mentions using embeddings in euclidean domain, later it performs 
label-smoothing in hyperbolic domain to refine the noisy encodings prior to the entity-typing. 
Performance evaluation using the benchmark datasets shows that the \FGETHR{} offers a perfect 
geometry for context sharing across distantly supervised data, and in turn outperforms the 
existing research on \FGNET{} by a significant margin.